\documentclass[10pt,twocolumn,letterpaper]{article}

\usepackage{cvpr}

\usepackage{epsfig}
\usepackage{graphicx}
\usepackage{amsmath}
\usepackage{amssymb}
\usepackage{stfloats}  \usepackage[flushleft]{threeparttable}  \usepackage{float}
\usepackage{dsfont}  \usepackage{enumitem}  \usepackage{xcolor}  \usepackage{multirow}  \usepackage{colortbl}
\definecolor{RED}{rgb}{1,0,0}
\definecolor{GRAY}{gray}{.9}
\def\ourmethod{{A$^2$U}\xspace}  \def\indexnet{{IndexNet}\xspace}  \def\carafe{{CARAFE}\xspace}

\usepackage{eucal}

\usepackage[pagebackref=true,breaklinks=true,letterpaper=true,colorlinks,bookmarks=false]{hyperref}

\usepackage[margin=0pt,font=small,labelfont=bf,labelsep=endash,tableposition=top]{caption}

 \graphicspath{{.}{./figures/}{../figures/}}

\cvprfinalcopy

\begin{document}

\title{Learning Affinity-Aware Upsampling for Deep Image Matting\thanks{HL's contribution was made when he was with The University of Adelaide.
Correspondence  should be addressed to CS.}
}

\author{Yutong Dai$^1$, ~~~~~~ Hao Lu$^2$, ~~~~~~ Chunhua Shen$^1$ \\
$^1$ The University of Adelaide, Australia  ~
$^2$ Huazhong University of Science and Technology, China
}

\maketitle

\begin{abstract}
We show that learning affinity in upsampling provides an effective and efficient
approach to
exploit
pairwise interactions in deep networks. Second-order features are commonly used in dense prediction to build adjacent relations with a learnable module after upsampling such as non-local blocks. Since upsampling is
essential,
learning affinity in upsampling can avoid additional propagation layers, offering the potential for building compact models. By looking
at
existing upsampling operators from a unified mathematical perspective, we generalize them into a second-order form and introduce Affinity-Aware Upsampling (\ourmethod) where upsampling kernels are generated
using
a light-weight low-rank bilinear model and are conditioned on second-order features.
Our upsampling operator can also be extended to downsampling. We discuss alternative
implementations of \ourmethod and verify their effectiveness on two detail-sensitive tasks:
image reconstruction
on a toy dataset;
and a large-scale image matting task where affinity-based ideas constitute
mainstream matting approaches.
In particular, results on the Composition-1k matting dataset show that
\ourmethod
achieves
a $14\%$ relative improvement in the SAD metric against
a strong
baseline with
negligible increase of
parameters ($< 0.5\%$). Compared with the state-of-the-art matting network, we achieve $8\%$ higher performance
with only $40\%$ model
complexity.
\end{abstract}

\section{Introduction}
\begin{figure}[t]
    \centering
    \includegraphics[width=\linewidth]{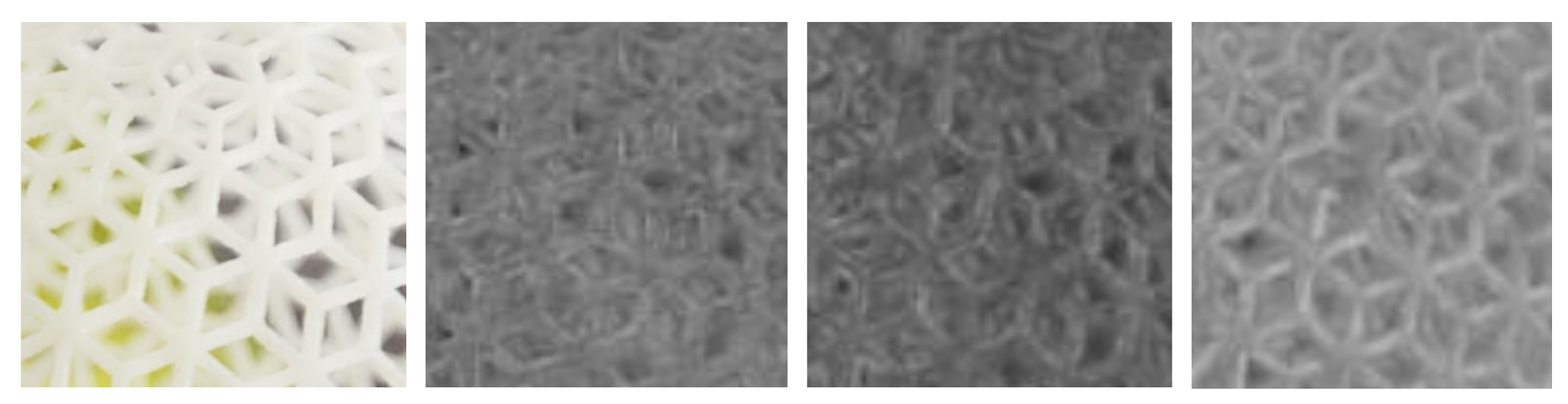}
    \caption{Visualization of upsampled feature maps with
various
    upsampling operators. From left to right, the input RGB image,
feature maps after the last upsampling
using nearest neighbor interpolation, bilinear upsampling, and our proposed affinity-aware upsampling, respectively. Our method produces better details with clear connectivity.}
    \label{fig:visualization}
    \vspace{-10pt}
\end{figure}

The similarity among positions, \textit{a.k.a.}\  affinity, is commonly investigated in dense prediction tasks~\cite{liu2017learning, cheng2018depth, gao2019ssap, wang2018deep,li2020natural}. Compared with directly fitting
ground truths using first-order features,
modeling similarity among different positions can provide second-order information.
There currently exist two solutions to learn affinity in deep networks: i) learning an affinity map before a non-deep backend and ii) defining a learnable affinity-based module to propagate information.
We are interested in end-to-end affinity learning, because classic methods often build upon some assumptions, rendering weak generalization in general cases.
Existing approaches typically propagate or model affinity after upsampling layers or before the last prediction layer. While affinity properties are modeled,
they sometimes may not be effective for the downstream tasks. For instance,the work in ~\cite{li2020natural} requires a feature encoding block besides the encoder-decoder architecture to learn affinity. The work in
\cite{cheng2018depth} needs more iterations to refine the feature maps according to their affinity at the last stage. As shown in Fig.~\ref{fig:visualization}, one plausible
reason is that pairwise similarity is
damaged
during upsampling. In addition, it is inefficient to construct interactions between high-dimensional feature maps. We therefore pose the question: \textit{Can we model affinity earlier in upsampling in an effective and efficient manner}?

Many widely used upsampling operators
interpolate values following a fixed
rule
at different positions. For instance, despite reference positions may change in bilinear upsampling, it always interpolates values
based on relative spatial distances. Recently,
the idea of learning to upsample emerges~\cite{lu2019indices, lu2020index,wang2019carafe}.
A learnable module is often built to generate upsampling kernels conditioned on feature maps to enable dynamic, feature-dependent upsampling behaviors.
Two such representative  operators include \carafe~\cite{wang2019carafe} and \indexnet~\cite{lu2020index}.
In our experiments, we find that \carafe may not work well in low-level vision tasks where details need to be restored.
\indexnet instead can recover details much better. We believe that one important reason is that \indexnet
encodes, stores, and delivers spatial information prior to downsampling. But computation can be costly when the network goes deep. This motivates us to pursue not only flexible but also light-weight designs of the upsampling operator.

In this paper, we propose to model affinity into upsampling and introduce a novel learnable upsampling operator, \textit{i.e.}, affinity-aware upsampling (\ourmethod). As we show later in Section~\ref{sec:mathmatical_perspective}, \textit{\ourmethod is a generalization of first-order upsampling operators}: in some conditions, the first-order formulation in~\cite{wang2019carafe} and~\cite{lu2019indices} can be viewed as special cases of our second-order one. In addition, by implementing \ourmethod in a low-rank bilinear formulation, we can achieve efficient upsampling with few extra parameters.

We demonstrate the effectiveness of \ourmethod on two detail-sensitive tasks:
an
image reconstruction task on a toy dataset with controllable background and a large-scale image matting task with subtle foregrounds. Image matting is a desirable task to justify the usefulness of affinity, because affinity-based matting approaches constitute one of prominent matting paradigms in
literatures.
Top
matting performance thus can suggest appropriate affinity modeling. In particular, we further discuss alternative design choices of \ourmethod and compare their similarities and differences. Compared with a strong image matting baseline on the Composition-1k matting dataset, \ourmethod exhibits a significant improvement ($\sim 14\%$) with
negligible
increase of parameters ($<0.5\%$),
proffering
a light-weight image matting architecture with state-of-the-art performance.

\section{Related work}
\noindent\textbf{Upsampling Operators in Deep Networks.}
Upsampling is often necessary in dense prediction
to recover spatial resolution.
The mostly used upsampling operators are bilinear interpolation and nearest neighbor interpolation. Since they are executed only based on spatial distances, they may be sub-optimal in detail-oriented tasks such as image matting where distance-based similarity can be violated.
Compared with distance-based upsampling, max-unpooling is feature-dependent
and has been shown to benefit
detail-oriented tasks~\cite{lu2019indices, lu2020index}, but it must
match
with max-pooling. In recent literatures, learning-based upsampling operators~\cite{shi2016real,long2015fully,wang2019carafe,lu2020index} emerge. The Pixel Shuffle (P.S.)~\cite{shi2016real} upsamples feature maps by reshaping. The deconvolution (Deconv)~\cite{long2015fully}, an inverse version of convolution, learns the upsampling kernel via back-propagation. Both P.S. and Deconv are data-independent during inference, because the kernel is fixed once learned.
By contrast, \carafe~\cite{wang2019carafe} and \indexnet~\cite{lu2019indices} learn the upsampling kernel dynamically conditioned on the data. They both introduce additional modules to learn upsampling kernels. Since the upsampling kernel is directly related to the feature maps, these upsampling operators are considered first-order.

Following the learning-based upsampling paradigm, we also intend to learn dynamic upsampling operators but to condition on second-order features to enable affinity-informed upsampling.
We show that, compared with first-order upsampling, affinity-informed upsampling not only achieves better performance but also introduces a light-weight learning paradigm.

\noindent\textbf{Deep Image Matting.}
Affinity dominates the majority of classic image matting approaches~\cite{levin2007closed,chen2013knn, chuang2001bayesian,he2011global}. The main assumption in propagation-based matting is that, similar alpha values can be propagated from known positions to unknown positions, conditioned on affinity. This assumption, however, highly depends on the color distribution. Such methods can perform well on cases with clear color contrast but more often fail in cases where the color distribution assumption is violated.
Recently, deep learning is found effective to address ill-posed image matting. Many deep matting methods arise~\cite{cho2016natural, wang2018deep, xu2017deep, tang2019learning, hou2019context, lu2019indices, li2020natural, cai2019disentangled}.
This field has experienced from a semi-deep stage~\cite{cho2016natural, wang2018deep} to a fully-deep stage~\cite{xu2017deep, hou2019context, lu2019indices, li2020natural, cai2019disentangled}. Here `semi-deep' means
that the matting part still relies on classic methods~\cite{levin2007closed,chen2013knn} to function,
while `fully-deep' means that the entire network does not resort to any classic algorithms. Among fully-deep matting, DeepMatting~\cite{xu2017deep} first applied the encoder-decoder architecture and reported
improved
results. Targeting this strong baseline, several deep matting methods were proposed. AlphaGAN matting~\cite{lutz2018alphagan} and IndexNet matting~\cite{lu2019indices} explored adversarial learning and index generating module to improve matting performance, respectively. In particular,
works in \cite{hou2019context, li2020natural, cai2019disentangled, tang2019learning} imitated classic sampling-based and propagation-based ideas into deep networks to ease the difficulty of learning. Therein, GCA matting~\cite{li2020natural} first designed an affinity-based module and demonstrated the effectiveness of affinity in fully-deep matting.
It treats alpha propagation as an independent module and adds it to different layers
to refine the feature map, layer by layer.

Different from the idea of `generating then refining', we propose to directly incorporate the propagation-based idea into upsampling for deep image matting. It not only benefits alpha propagation but also shows the potential for light-weight module design.

\section{A Mathematical View of Upsampling}
\label{sec:mathmatical_perspective}

The work in
\cite{lu2020index} unifies upsampling from an indexing perspective. Here we provide an alternative mathematical view.
To simplify exposition, we discuss the upsampling of the one-channel feature map. Without loss of generality, the one-channel case can be easily extended to multi-channel upsampling, because most upsampling operators execute per-channel upsampling. Given a one-channel local feature map $\mathbf{Z}\in\mathbb{R}^{k\times k}$ used to generate an upsampled feature point, it can be vectorized to $\mathbf{z}\in\mathbb{R}^{k^2\times1}$. Similarly, the vectorization of an upsampling kernel $\mathbf{W}\in\mathbb{R}^{k\times k}$ can be denoted by $\mathbf{w}\in\mathbb{R}^{k^2\times1}$.
If ${\mathit{g(\mathbf{w}, \mathbf{z})}}$ defines the output of upsampling, most existing upsampling operations follow
\begin{equation}\label{eq:general_formulation}
    {\mathit{g(\mathbf{w}, \mathbf{z})}}= {\mathbf{w}}^{T}{\mathbf{z}}\,.
\end{equation}
Note that ${\mathit{g(\mathbf{w}, \mathbf{z})}}$ indicates an upsampled point. In practice, multiple such points can be generated to form an upsampled feature map.
$\mathbf{w}$ may be
either
shared or unshared among channels depending on the upsampling operator. Different operators define different ${\mathbf{w}}$'s.
Further, even the same ${\mathbf{w}}$ can be applied to different ${\mathbf{z}}$'s.
According to how the upsampling kernel $\mathbf{w}$ is generated, we categorize the kernel into two types: the universal kernel and the customized kernel. The universal kernel is input-independent. It follows the same upsampling rule given any input. One example is deconvolution~\cite{long2015fully}.
The customized kernel, however,
is input-dependent. Based on what input is used to generate the kernel, the customized kernel can be further divided into distance-based and feature-based. We elaborate as
follows.

\vspace{5pt}
\noindent\textbf{Distance-based Upsampling.}
Distance-based upsampling is implemented according to spatial distances, such as nearest neighbor and bilinear interpolation. The difference between them is the number of positions taken into account. Under the definition of Eq.~\eqref{eq:general_formulation}, the upsampling kernel is a function of the relative distance between points. By taking bilinear interpolation with $4$ reference points as an example,
${\mathbf{w}}=[w_{1}, w_{2}, w_{3}, w_{4}]$,
where ${ w }_{ 1 }=\frac { 1 }{ \left( { x }_{ 1 }-{ x }_{ 0 } \right) \left( { y }_{ 1 }-{ y }_{ 0 } \right)  } \left( { x }_{ 1 }-x \right) \left( { y }_{ 1 }-y \right) $ given the coordinates of two reference points $(x_0, y_0)$ and $(x_1, y_1)$; $x$ and $y$ is the coordinates of the interpolated point; $w_{2}$, $w_{3}$, and $w_{4}$ can be derived similarly. In multi-channel cases, the same ${\mathbf{w}}$ is shared by all channels of input.

\vspace{5pt}
\noindent\textbf{Feature-based Upsampling.}
Feature-based upsampling is feature-dependent. They are developed in deep networks, including max-unpooling~\cite{badrinarayanan2017segnet}, \carafe~\cite{wang2019carafe}, and \indexnet~\cite{lu2020index}:
\begin{enumerate}[label=\roman*), leftmargin=1.1em]
\itemsep -0.1cm
    \item \textit{Max-unpooling} interpolates values following the indices returned from max-pooling. In a $2\times2$ region of the feature layer after upsampling, only one position recorded in the indices has value, and other three are filled with $0$. Since each position on the upsampled feature map is interpolated from a $1\times 1$ point at the low-resolution layer, we can define $\mathbf{w}$ by a $1\times 1$ vector ${\mathbf{w}}=[w]$, where $w\in\mathbb{R}^{1\times1}$, and $\mathbf{z}$ is also the $1\times 1$ point at the low-resolution layer. Note that, $w\in\{0,1\}$, and only one $w$ can equal to $1$ in a $2\times2$ region of the output feature map. In multi-channel cases, ${\mathbf{w}}$ and ${\mathbf{z}}$ are different in different channels conditioned on the $\max$ operator.

    \item \textit{\carafe} learns an upsampling kernel $\mathbf{w}\in \mathbb{R}^{k^2\times1}$ ($k=5$ in~\cite{wang2019carafe}) via a kernel generation module given a decoder feature map ready to upsample. It also conforms to Eq.~\eqref{eq:general_formulation}, where $\mathbf{z}\in \mathbb{R}^{k^2\times1}$ is obtained from the low-resolution decoder feature map. The kernel size of $\mathbf{w}$ depends on the size of $\mathbf{z}$. In multi-channel cases, the same ${\mathbf{w}}$ is shared among channels.

    \item \textit{\indexnet} also learns an upsampling kernel dynamically from features. The difference is that \indexnet learns from high-resolution encoder feature maps. Under the formulation of Eq.~\eqref{eq:general_formulation}, the upsampling kernel follows a similar spirit like max-unpooling: ${\mathbf{w}}=[w]$, where $w\in\mathbb{R}^{1\times1}$, because each position on the upsampled feature layer is interpolated from a corresponding point on the low-resolution map by multiplying by an interpolation weight $w$. But here $w\in[0,1]$ instead of $\{0,1\}$.
\end{enumerate}

Hence, distanced-based and feature-based upsampling operators have a unified form ${\mathit{g(\mathbf{w}, \mathbf{z})}}={\mathbf{w}}^{T}{\mathbf{z}}$, while different operators
correspond to
different $\mathbf{w}$'s and $\mathbf{z}$'s, where $\mathbf{w}$ can be heuristically defined or dynamically generated. In particular, existing operators define/generate $\mathbf{w}$ according to distances or first-order features, while \textit{second-order information remains unexplored in upsampling}.

\section{Learning Affinity-Aware Upsampling}
Here we explain
how we exploit second-order information to formulate the affinity idea
in
upsampling using a bilinear model and how we apply a low-rank approximation to reduce computational complexity.

\vspace{5pt}
\noindent\textbf{General Formulation of Upsampling.}
Given a feature map $\mathcal{M}\in { \mathbb{R} }^{C\times H\times W }$ to be upsampled, the goal is to generate an upsampled feature map $\mathcal{M}'\in { \mathbb{R} }^{C\times rH\times rW }$, where $r$ is the upsampling ratio.
For a position $(i',j')$ in $\mathcal{M}'$, the corresponding source position $(i,j)$ in $\mathcal{M}$ is derived by solving $i=\lfloor i'/r\rfloor$, $j=\lfloor j'/r\rfloor$.
We aim to learn an upsampling kernel $\mathbf{ w }\in { \mathbb{R} }^{k^2\times1}$ for each position in $\mathcal{M}'$.
By applying the kernel to a channel of the local feature map $\mathcal{X}\in \mathbb{R}^{C\times k\times k}$ centered at position $l$ on $\mathcal{M}$, denoted by $\mathbf{X}\in\mathbb{R}^{1\times k\times k}$,
the corresponding upsampled feature point $m_{l'}'\in\mathcal{M}'$ of the same channel at target position $l'$ can be obtained by $m_{l'}'=\mathbf{w}^T\mathbf{x}$ according to Eq.~\eqref{eq:general_formulation}, where $\mathbf{x}\in\mathbb{R}^{k^2\times1}$ is the vectorization of $\mathbf{X}$.

\vspace{5pt}
\noindent\textbf{General
Meaning
of Affinity.} Affinity is often used to indicate pairwise similarity and is considered second-order features. An affinity map can be constructed in different ways such as using a Gaussian kernel.
In self-attention, the affinity between the position $l$ and the enumeration of all possible positions $p$ at a feature map $\mathcal{M}$ is denoted by $\underset { \forall p}{ softmax } \left( sim\left( \mathbf{m}_{ l },\mathbf{ m }_{ p  } \right)  \right) $, where $\mathbf{m}_{l}$ and $\mathbf{ m }_{   }$ represent two vectors at position $l$ and $p$, respectively, and $sim\left( \mathbf{m}_{ l },\mathbf{ m }_{ p } \right)$ measures the similarity between $\mathbf{m}_{l}$ and $\mathbf{m}_{p }$ with the inner product ${\mathbf{m}_{ l}}^{T}\mathbf{m}_{ p}$.

\vspace{5pt}
\noindent\textbf{Affinity-Aware Upsampling via Bilinear Modeling.}
Given a local feature map $\mathcal{X}\in \mathds{ R }^{ C\times { h }_{ 1 }\times { w }_{ 1 } }$, $\mathcal{X}$ has an equivalent matrix form $\mathbf{X}\in \mathbb{R}^{C\times N}$, where $N=h_{1}\times w_{1}$. We aim to learn an upsampling kernel conditioned on $\mathbf{X}$. Previous learning-based upsampling operators~\cite{wang2019carafe, lu2019indices, lu2020index} generate the value of the upsampling kernel following a linear model by $w=\sum _{ i=1 }^{ C }\sum _{ j=1 }^{N }{ { { a }_{ ij }{x} }_{ ij } } $, where $a_{ij}$ and ${x}_{ij}$ are the weight and the feature at the channel $i$ and position $j$ of $\mathbf{X}$, respectively. Note that $w\in\mathbb{R}^{1\times1}$. To encode second-order information, a natural generalization of the linear model above is bilinear modeling where another feature matrix $\mathbf{Y}\in \mathbb{R}^{C\times M}$
transformed from the feature map $\mathcal{Y}\in \mathds{ R }^{ C\times { h }_{ 2 }\times { w }_{ 2 } }$
($M=h_{2}\times w_{2}$), is introduced to pair with $\mathcal{X}$ to model affinity. Given each $\mathbf{x}_i\in\mathbb{R}^{C\times1}$ in $\mathbf{X}$, $\mathbf{y}_j\in\mathbb{R}^{C\times1}$ in $\mathbf{Y}$, the bilinear weight $a_{ij}$ of the vector pair, and the embedding weights $q_k$ and $t_k$ for each channel of $\mathbf{x}_{i}$ and $\mathbf{y}_{j}$, we propose to generate each value of the upsampling kernel from embedded pairwise similarity, \textit{i.e.},
\begin{equation}
\label{bilinear_model}
\begin{aligned}
   w &=\sum _{ i=1 }^{ N }{ \sum _{ j=1 }^{ M }{ a_{ ij }{ \varphi({\mathbf{ x } }_{ i } )}^{ T } {\phi(\mathbf{ y } }_{ j }) }  } =\sum _{ k=1 }^{ C } \sum _{ i=1 }^{ N }{ \sum _{ j=1 }^{ M }{ a_{ ij } }{q_{k}}{ x_{ ik } }{t_{k}}{ { y }_{ jk } } } \\
   & =\sum _{ k=1 }^{ C } \sum _{ i=1 }^{ N }{ \sum _{ j=1 }^{ M }{ a_{ ijk }' }{x_{ ik } }{ { y }_{ jk } } }=\sum _{ k=1 }^{ C }{ { {\mathbf{ x }} _{ k } }^{ T }{ \mathbf{ A }_{ k } }{ {\mathbf{ y }} _{ k } } }  \,,
\end{aligned}
\end{equation}
where $\mathbf{x}_k\in\mathbb{R}^{N\times1}$ and $\mathbf{y}_{k}\in\mathbb{R}^{M\times1}$ are the $k$-th channel of $\mathbf{X}$ and $\mathbf{Y}$, respectively, $\mathbf{A}_{k}\in\mathbb{R}^{N\times M}$ is the affinity matrix for $k$-th channel, $a_{ijk}'=a_{ij}q_{k}t_{k}$, and $\varphi$ and $\phi$ represent the embedding function.

\begin{figure*}[!t]
    \centering
    \includegraphics[width=.78807\linewidth]{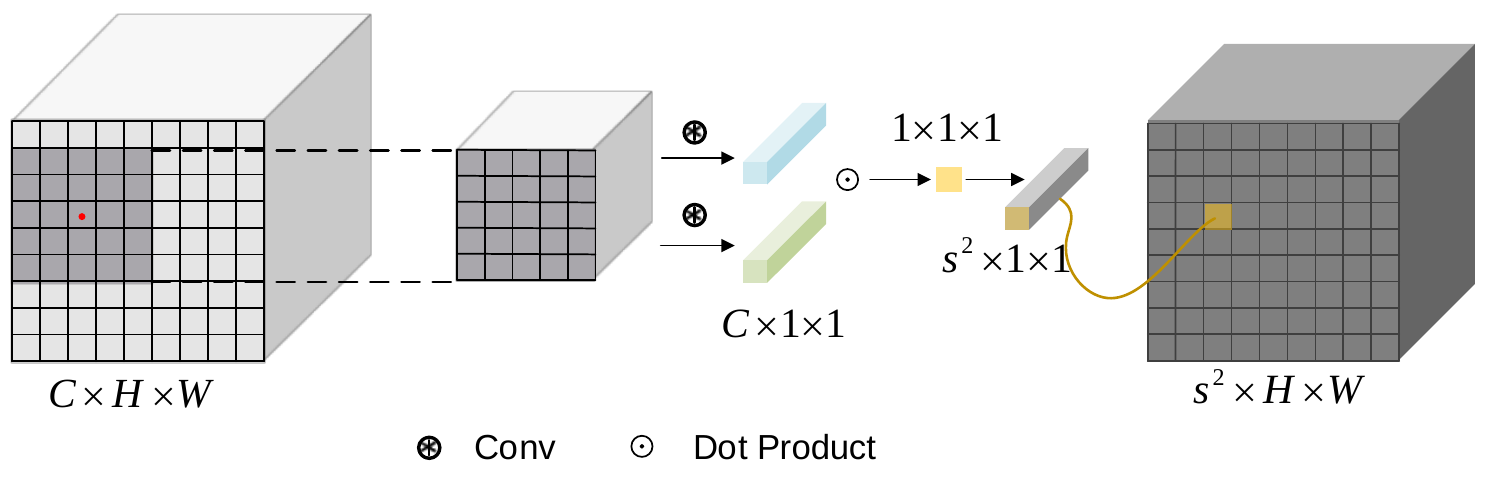}\vspace{-10pt}
    \caption{Kernel generation of \ourmethod. Given a feature map of size $C\times H\times W$, a $s\times s$ upsampling kernel is generated at each spatial position conditioned on the feature map. The rank $d$ is $1$ here.}
    \label{fig:upsampling_operator}
\end{figure*}

\vspace{5pt}
\noindent\textbf{Factorized Affinity-Aware Upsampling.}
Learning $\mathbf{A}_{k}$ can be expensive when $M$ and $N$ are large.
Inspired by~\cite{kim2016hadamard,yu2018hierarchical}, a low-rank bilinear method can be derived to reduce computational complexity of Eq.~\eqref{bilinear_model}. Specifically, $\mathbf{A} _{k}$ can be rewritten by $\mathbf{A} _{k}=\mathbf{U}_{k}\mathbf{V}_{k}^{T}$, where $\mathbf{U}_{k}\in{\mathbb{R}}^{N\times d}$ and $\mathbf{V}_{k}\in{\mathbb{R}}^{M\times d}$. $d$ represents the rank of $\mathbf{A}_{k}$ under the constraint of $d\le \min(N,M)$. Eq.~\eqref{bilinear_model} therefore can be rewritten by
\begin{equation} \label{factorized_bilinear}
\begin{aligned}
    w&=\sum_{k=1}^{C}{\mathbf{x}_{k}}^{T}{\mathbf{U}}_{k}{{\mathbf{V}}_{k}}^{T}{\mathbf{y}_{k}}=\sum_{k=1}^{C}{\mathds{1}}^{T}({{\mathbf{U}}_{k}}^{T}{\mathbf{x}_k}\circ{{\mathbf{V}}_{k}}^{T}{\mathbf{y}_k}) \\
    &={\mathds{1}}^{T}\sum_{k=1}^{C}({{\mathbf{U}}_{k}}^{T}{\mathbf{x}_{k}}\circ{{\mathbf{V}}_{k}}^{T}{\mathbf{y}_{k}})
\end{aligned}\,,
\end{equation}
where $\mathds{1}\in\mathbb{R}^{d}$ is a column vector of ones, and $\circ$ denotes the Hadamard product.
Since we need to generate a $s\times s$ upsampling kernel, $\mathds{1}$ in Eq.~\eqref{factorized_bilinear} can be replaced with ${\mathbf{P}}\in \mathbb{R}^{d\times s^2}$. Note that, Eq.~\eqref{factorized_bilinear} is applied to each position of a feature map, so the inner product here can be implemented by convolution. The full upsampling kernel therefore can be generated by
\begin{equation} \label{final_bilinear}
\begin{aligned}
\mathbf{w}&={\mathbf{P}}^{T}\sum_{k=1}^{C}({{\mathbf{U}}_{k}}^{T}{\mathbf{x}_{k}}\circ{{\mathbf{V}}_{k}}^{T}{\mathbf{y}_{k}}) \\
    &={\mathbf{P}}^{T}\overset { d }{ \underset { r=1 }{\tt {cat}}}\Big(\sum_{k=1}^{C}({{\mathbf{u}}_{kr}}^{T}{\mathbf{x}_{k}}\circ{{\mathbf{v}}_{kr}}^{T}{\mathbf{y}_{k}})\Big) \\
    &={\tt{conv}}\Big(\mathcal{P}, \overset { d }{ \underset { r=1 }{\tt{cat}}}\big({\tt{gpconv}}(\mathcal{U}_{r}, \mathcal{X})\odot {\tt{gpconv}}(\mathcal{V}_{r}, \mathcal{Y})\big)\Big)
\end{aligned}\,,
\end{equation}
where $\mathbf{u}_{kr}\in\mathbb{R}^{N\times 1}$, $\mathbf{v}_{kr}\in\mathbb{R}^{M\times 1}$. The convolution kernels $\mathcal{P}\in\mathbb{R}^{d\times s^2\times1\times 1}$, $\mathcal{U}\in\mathbb{R}^{d\times C\times h_{1}\times w_{1}}$, and $\mathcal{V}\in\mathbb{R}^{d\times C\times h_{2}\times w_{2}}$ are reshaped tensor versions of $\mathbf{P}$, $\mathbf{U}$ and $\mathbf{V}$, respectively. $\tt{conv}(\mathcal{K}, \mathcal{M})$ represents a convolution operation on the feature map $\mathcal{M}$ with the kernel $\mathcal{K}$; $\tt{gpconv}(\mathcal{K}, \mathcal{M})$ defines a group convolution operation ($C$ groups) with the same input.
$\tt{cat}$ is the concatenate operator.
This process is visualized in Fig.~\ref{fig:upsampling_operator}.

\vspace{5pt}
\noindent\textbf{Alternative Implementations.}
Eq.~\eqref{final_bilinear} is a generic formulation. In practice, many design choices can be discussed in implementation:
\begin{enumerate}[leftmargin=1.1em,label=\roman*)]
\itemsep -0.1cm
    \item The selection of $\mathcal{X}$ and $\mathcal{Y}$ can be either same or different. In this paper, we only discuss self-similarity, i.e., $\mathcal{X}=\mathcal{Y}$;

    \item The rank $d$ can be chosen in the range $[1, \min(N,M)]$. For example, if $\mathcal{X}$ and $\mathcal{Y}$ are extracted in $5\times 5$ regions, the range will be $[1, 25]$. In our experiments, we set $d=1$ to explore the most simplified and light-weight case.

    \item $\mathcal{U}$ and $\mathcal{V}$ can be considered two encoding functions. They can be shared, partly-shared, or unshared among channels. We discuss two extreme cases in the experiments: `channel-shared' (`\textit{cs}') and `channel-wise' (`\textit{cw}').

    \item Eq.~\eqref{final_bilinear} adjusts the kernel size of $\mathbf{w}$ only using $\mathcal{P}$. Since the low-rank approximation has less parameters, fixed $\mathcal{P}$, $\mathcal{U}$, and $\mathcal{V}$ may not be sufficient to model all local variations. Inspired by CondConv~\cite{yang2019condconv}, we attempt to generate $\mathcal{P}$ and $\mathcal{U}$, $\mathcal{V}$ dynamically conditioned on the input. We investigate three implementations: 1) \textit{static}: none of them is input-dependent; 2) \textit{hybrid}: only $\mathcal{P}$ is conditioned on input; and 3) \textit{dynamic}: $\mathcal{P}$, $\mathcal{U}$, and $\mathcal{V}$ are all conditioned on input. The dynamic generation of $\mathcal{P}$, $\mathcal{U}$, or $\mathcal{V}$ is implemented using a global average pooling and a $1\times1$ convolution layer.

    \item We implement stride-2  $\mathcal{U}$ and $\mathcal{V}$ in our experiments. They output features of size $C\times \frac{H}{2}\times \frac{W}{2}$. To generate an upsampling kernel of size $s^2\times H\times W$, one can either use $4$ sets of different weights for $\mathcal{U}$ and $\mathcal{V}$ or $4$ sets of weights for $\mathcal{P}$ ($4\times s^2\times \frac{H}{2}\times \frac{W}{2}$), followed by a shuffling operation ($s^2\times H\times W$). We denote the former case as `pointwise' (`\textit{pw}'). Further, as pointed
    out
    in~\cite{kim2016hadamard}, nonlinearity, \textit{e.g.},
    \texttt{tanh} or \texttt{relu}, can be added after the encoding of $\mathcal{U}$ and $\mathcal{V}$. We verify a similar idea by adding normalization and nonlinearity in the experiments.

\end{enumerate}

\begin{table*}[!ht] \small
    \centering
    \addtolength{\tabcolsep}{1.5pt}
    \begin{tabular}{l | c c c c|c c c c}
    \hline
   Method  & \multicolumn{4}{c|}{MNIST} & \multicolumn{4}{c}{Fashion-MNIST} \\
& PSNR ($\uparrow$) & SSIM ($\uparrow$) & MSE ($\downarrow$) & MAE ($\downarrow$) & PSNR ($\uparrow$) & SSIM ($\uparrow$) & MSE ($\downarrow$) & MAE ($\downarrow$) \\
         \hline
     Conv\textsubscript{/2}-Nearest & 28.54 & 0.9874 & 0.0374 & 0.0148 & 25.58 & 0.9797 & 0.0527 & 0.0269 \\
     Conv\textsubscript{/2}-Bilinear & 26.12 & 0.9783 & 0.0495 & 0.0205 & 23.68 & 0.9675 & 0.0656 & 0.0343 \\
     Conv\textsubscript{/2}-Deconv~\cite{long2015fully}  & 31.85 & 0.9942 & 0.0256 & 0.0089 & 27.42 & 0.9870 & 0.0426 & 0.0207 \\
     P.S.~\cite{shi2016real}  & 31.63 & 0.9939 & 0.0262 & 0.0099 & 27.33 & 0.9868 & 0.0431 & 0.0212 \\
MaxPool-MaxUnpool  & 29.91 & 0.9916 & 0.0320 & 0.0133 & 28.31 & 0.9901 & 0.0385 & 0.0218 \\
     MaxPool-\carafe~\cite{wang2019carafe} & 28.72 & 0.9885 & 0.0367 & 0.0131 & 25.17 & 0.9773 & 0.0552 & 0.0266 \\
     MaxPool-\indexnet\textsuperscript{$\dagger$}~\cite{lu2019indices}  & 45.51 & 0.9997 & 0.0053 & 0.0024 & 45.83 & 0.9998 & 0.0051 & 0.0033 \\
MaxPool-\ourmethod (Ours)  & 47.63 & 0.9998 & 0.0042 & 0.0020 & 46.41 & 0.9999 & 0.0048 & 0.0031 \\
\hline
     MaxPool-\indexnet\textsuperscript{$\ddagger$}~\cite{lu2019indices}  & 47.13 & 0.9997 & 0.0044 & 0.0020 & 44.35 & 0.9998 & 0.0061 & 0.0036 \\
\hline
    \end{tabular}

\caption{Reconstruction results on the MNIST dataset and the Fashion-MNIST dataset. $^\dagger$ denotes holistic index network, $^\ddagger$ represents depthwise index network. Both index networks here apply the setting of `context+linear' for a fair comparison.}
    \label{tab:toy_experiment}
    \vspace{-15pt}
\end{table*}
\vspace{5pt}
\noindent\textbf{Extension to Downsampling.}
Following~\cite{lu2020index}, our method can also be extended to downsampling. Downsampling is in pair with upsampling, so their kernels are generated from the same encoder feature. We use `\textit{d}' to indicate the use of paired downsampling in experiments. We share the same $\mathcal{U}$ and $\mathcal{V}$ in Eq.~\eqref{final_bilinear} in both downsampling and upsampling, but use different $\mathcal{P}$'s considering that they may have different kernel sizes. We denote the overall upsampling kernel by $\mathcal{W}_{u}\in{\mathbb{R}^{{s_{u}}^2\times H \times W}}$ and the downsampling kernel by $\mathcal{W}_{d}\in{\mathbb{R}^{{s_{d}}^2\times H/r \times W/r}}$, where $r$ is the ratio of upsampling/downsampling. We set $s_{d}=r^{2}s_{u}$ in our experiments.

\section{Image Reconstruction and Analysis}
Here we
conduct a
pilot image reconstruction experiment
on a toy dataset
to show the effectiveness of \ourmethod. Inspired by~\cite{lu2020index}, we build sets of reconstruction experiments on the MNIST dataset~\cite{lecun1998mnist} and Fashion-MNIST dataset~\cite{xiao2017fashion}. The motivation is to verify whether
exploiting second-order information into upsampling benefits recovering spatial information.

We denote $C(k)$ to be a convolution layer with $k$-channel output and $3\times3$ filters (stride is $1$ unless stated), followed by BatchNorm and ReLU, and denote $D_{r}$ a downsampling operator with a ratio of $r$, and denote $U_{r}$ an upsampling operator with a ratio of $r$. We build the network architecture as: $C(32)$-$D_{2}$-$C(64)$-$D_{2}$-$C(128)$-$D_{2}$-$C(256)$-$C(128)$-$U_{2}$-$C(64)$-$U_{2}$-$C(32)$-$U_{2}$-$C(1)$. The same training strategies and evaluation metrics are used following~\cite{lu2020index}. Since training patches are relatively small ($32\times32$), upsampling kernel sizes for \carafe and \ourmethod are both set to $1$, and the encoding convolution kernels in \indexnet and \ourmethod are both set to $4$. Other settings keep the default ones. We apply `static-pw-cw' \ourmethod here because it is the same as Holistic IndexNet if letting convolution results of $\mathcal{U}$ to be all ones. We hence add a \texttt{sigmoid} function after $\mathcal{U}$ to generalize IndexNet. To avoid extra layers, we apply max-pooling to $D_{r}$ to obtain high-resolution layers when validating \indexnet and \ourmethod. Reconstruction results are presented in Table~\ref{tab:toy_experiment}.

As shown in Table~\ref{tab:toy_experiment}, upsampling operators informed by features (max-unpooling,  \carafe, \indexnet, and \ourmethod) outperform the operators guided by spatial distances (nearest, bilinear, and bicubic). Moreover,
learning from high-resolution features matter for upsampling, among which, learning-based operators (\indexnet, \ourmethod) achieve the best results. Further, it is worth noting that, \ourmethod performs better than \indexnet with even fewer parameters. From these observations, we believe in upsampling: 1) high-resolution features are beneficial to extract spatial information, and 2) second-order features can help to recover more spatial details than first-order ones.

\section{Experiments and Discussions}
Here we evaluate \ourmethod on deep image matting. This task is suitable for assessing the quality of modeling pairwise relations.

\begin{figure*}[t]
    \centering
    \includegraphics[width=0.98\linewidth]{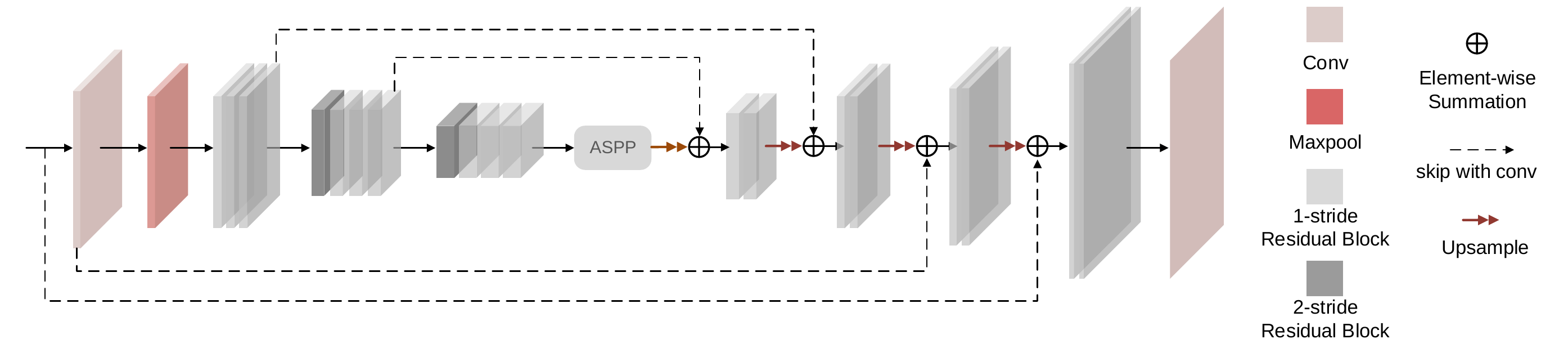}
    \caption{Overview of our matting framework. The focus of this work is on the upsampling stages.
}
    \vspace{-10pt}
    \label{fig:baseline_architecture}
\end{figure*}
\subsection{Network Architecture}
Similar to~\cite{li2020natural}, our baseline network adopts the first $11$ layers of the ResNet34~\cite{he2016deep} as the encoder. The decoder consists of residual blocks and upsampling stages. The In-Place Activated BatchNorm~\cite{rotabulo2017place} is applied to each layer except the last one to reduce GPU memory consumption. As shown in Fig.~\ref{fig:baseline_architecture}, the overall network follows the UNet architecture~\cite{ronneberger2015u} with `skip' connection. To apply \ourmethod to upsampling, we replace the upsampling operations in the decoder with \ourmethod modules. Specifically, we learn upsampling kernels from the skipped features. If \ourmethod is used in both upsampling and downsampling stages, we change all 2-stride convolution layers in the encoder to be 1-stride and implement paired downsampling and upsampling operations,
respectively, by learning upsampling/downsampling kernels from the modified 1-stride feature layer.

\subsection{Datasets}
We mainly conduct our experiments on the Adobe Image Matting dataset~\cite{xu2017deep}. Its training set has $431$ unique foreground objects and ground-truth alpha mattes. Instead of compositing each foreground with fixed $100$ background images chosen from MS COCO~\cite{lin2014microsoft}, we randomly choose the background images in each iteration and generate the composition images on-the-fly. The test set, termed the Composition-1k, contains $50$ unique foreground objects; each foreground is composited with $20$ background images from the Pascal VOC dataset~\cite{everingham2010pascal}.

We also evaluate our method on the $\tt alphamatting.com$ benchmark~\cite{rhemann2009perceptually}. This online benchmark has $8$ unique testing images and $3$ different trimaps for each image, providing $24$ test cases.

Further, we report results on the recently proposed Distinctions-646 dataset~\cite{qiao2020attention}. It has $596$ foreground objects in the training set and $50$ foreground objects in the test set. We generate the training data and the test set following the same protocol as on the Adode Image Matting dataset.

\subsection{Implementation Details}
Our implementation is based on PyTorch~\cite{paszke2019pytorch}. Here we describe training details on the Adobe Image Matting dataset. The $4$-channel input concatenates the RGB image and its trimap. We mainly follow the data argumentation of~\cite{li2020natural}. Two foreground objects are first chosen with a probability of $0.5$ and are composited to generate a new foreground image and a new alpha matte. Next, they are resized to $640\times 640$ with a probability of $0.25$. Random affine transformations are then applied. Trimaps are randomly dilated from the ground truth alpha mattes with distances in the range between $1$ and $29$, followed by $512\times 512$ random cropping.
The background image is randomly chosen from the MS COCO dataset~\cite{lin2014microsoft}. After imposing random jitters to the foreground object, the RGB image is finally generated by composition.

The backbone is pretrained on ImageNet~\cite{krizhevsky2017imagenet}. Adam optimizer~\cite{kingma2014adam} is used. We use the same loss function as~\cite{xu2017deep, lu2019indices}, including alpha prediction loss and composition loss computed from the unknown regions indicated by trimaps. We update parameters for $30$ epochs. Each epoch has a fixed number of $6000$ iterations. A batch size of $16$ is used and BN layers in the backbone are fixed. The learning rate is initialized to $0.01$ and reduced by $\times 10$ at the $20$-th epoch and the $26$-th epoch, respectively. The training strategies on the Distinction646 dataset are
the same except that we update the parameters for only $25$ epochs. We evaluate our results using Sum of Absolute Differences (SAD), Mean Squared Error (MSE), Gradient (Grad), and Connectivity (Conn)~\cite{rhemann2009perceptually}. We follow the evaluation code provided by~\cite{xu2017deep}.

\subsection{The Adobe Image Matting Dataset}
\begin{table}[!t] \footnotesize
    \centering
    \addtolength{\tabcolsep}{-1pt}
    \begin{tabular}{l c c c c r}
    \hline
        Upsample & SAD & MSE & Grad & Conn & \# Params \\
        \hline
        Nearest & 37.51 & 0.0096 & 19.07 & 35.72 & 8.05M \\
        Bilinear & 37.31 & 0.0103 & 21.38 & 35.39 & 8.05M \\
        \carafe & 41.01 & 0.0118 & 21.39 & 39.01 & +0.26M \\
        \indexnet & 34.28 & 0.0081 & \textbf{15.94} & 31.91 & +12.26M \\
        \hline
        \ourmethod (static-pw-cw)  & 36.36 & 0.0099 & 21.03 & 34.40 & +0.10M  \\
        \ourmethod (static-cw)  & 35.92 & 0.0098 & 20.06 & 33.68 & +26K \\
        \ourmethod (hybrid-cw)  & 34.76 & 0.0088 & 16.39 & 32.29 & +44K \\
        \ourmethod (hybrid-cs)  & 36.43 & 0.0098 & 21.24 &  34.11 & +19K \\
         \ourmethod (dynamic-cw) & 36.66 & 0.0094 & 18.60 & 34.62 & +0.20M \\
         \ourmethod (dynamic-cs) & 35.86 & 0.0095 & 17.13 & 33.71 & +20K \\
        \ourmethod (dynamic-cs-d) & 33.13 & \textbf{0.0078} & 17.90 & 30.22 & +38K \\
        \ourmethod (dynamic-cs-d)\textsuperscript{$\dagger$} & \textbf{32.15} & 0.0082 & 16.39 & \textbf{29.25} & +38K \\
        \hline
    \end{tabular}
\caption{Results of different upsampling operators on the Composition-1k test set with the same baseline model. $^\dagger$ denotes additional normalization and nonlinearity after the encoding layers of $\mathcal{U}$ and $\mathcal{V}$. The best performance is in boldface.}
\label{tab:adobe_upsample}
    \vspace{-15pt}
\end{table}

\noindent\textbf{Ablation Study on Alternative Implementations.}
Here we verify different implementations of \ourmethod on the Composition-1k test set and compare them with existing upsampling operators. Quantitative results are shown in Table~\ref{tab:adobe_upsample}. All the models
are implemented by the same architecture but with different upsampling operators. The `nearest' and `bilinear' are our direct baselines. They achieve close performance with the same model capacity. For \carafe, we use the default setting as in~\cite{wang2019carafe}, \textit{i.e.}, $k_{up}=5$ and $k_{encoder}=3$. We observe \carafe has a negative effect on the performance. The idea behind \carafe is to reassemble contextual information, which is not the focus of matting where subtle details matter. However, it is interesting that \carafe can still be useful for matting when it follows a light-weight MobileNetV2 backbone~\cite{lu2020index}.
One possible explanation is that a better backbone (ResNet34) suppresses the advantages of context reassembling. We report results of IndexNet with the best-performance setting (`depthwise+context+nonlinear') in~\cite{lu2019indices,lu2020index}. The upsampling indices are learned from the skipped feature layers. IndexNet achieves a notable improvement, especially on the Grad metric. However, IndexNet significantly increases the number of parameters.

We further investigate $6$ different implementations of \ourmethod and another version with paired downsampling and upsampling. According to the results, the `static' setting can only improve the SAD and Conn metrics. The position-wise and position-shared settings report comparable results, so we fix the position-shared setting in the following `hybrid' and `dynamic' experiments. We verify both channel-wise and channel-shared settings for `hybrid' and `dynamic' models. The `hybrid' achieves higher performance with channel-wise design, while the `dynamic' performs better with channel-shared design. All `hybrid' and `dynamic' models show improvements against baselines on all metrics, except the MSE and Grad metrics for the channel-shared `hybrid' model. The last implementation, where channel-shared `dynamic' downsampling is paired with upsampling, achieves the best performance (at least $14\%$ relative improvements against the baseline) with negligible
increase of parameters ($<0.5\%$).

Hence, while the dedicated design of upsampling operators matters, paired downsampling and upsampling seems more important, at least for image matting.

\begin{table}[!t] \footnotesize
    \centering
    \addtolength{\tabcolsep}{1.25pt}
    \begin{tabular}{l c c c c c c}
    \hline
    Method & $k_{up}$ & SAD & MSE & Grad & Conn \\
    \hline
    \ourmethod (hybrid-cw)  & 1 & 37.74 & 0.0104 & 22.07 & 35.91 \\
    \ourmethod (hybrid-cw) & 3 & 34.76 & 0.0088 & 16.39 & 32.29 \\
    \ourmethod (hybrid-cw) & 5 & 35.99 & 0.0093 & 17.96 & 33.90 \\
    \ourmethod (dynamic-cs) & 1 & 36.06 & 0.0098 & 17.25 & 33.95 \\
    \ourmethod (dynamic-cs) & 3 & 35.86 & 0.0095 & 17.13 & 33.71 \\
    \ourmethod (dynamic-cs) & 5 & 37.40 & 0.0096 & 18.28 & 35.50 \\
    \hline
    \end{tabular}
\caption{Ablation study of upsampling kernel size on the Composition-1k test set.}
    \label{tab:upsample_kernelsize}
    \vspace{-20pt}
\end{table} \vspace{5pt}
\noindent\textbf{Ablation Study on Upsampling Kernel.}
Here we investigate the performance of our models with different upsampling kernel sizes. The encoding kernel size (the kernel size of $\mathcal{U}$ or $\mathcal{V}$) is set to $k_{en}=5$ in all matting experiments unless stated. Under this setting, results in Table~\ref{tab:upsample_kernelsize} show that $k_{up}=3$ performs the best. It is interesting to observe that larger upsampling kernel does not imply better performance. We believe this is related to the encoding kernel size and the way how we generate $\mathcal{U}$, $\mathcal{V}$ and $\mathcal{P}$.
We use $k_{up}=3$ as our default setting.

\vspace{5pt}
\noindent\textbf{Ablation Study on Normalization.}
In both~\cite{wang2019carafe} and~\cite{lu2020index}, different normalization strategies are verified, and experiments show that normalization significantly affects the results. We thus justify the normalization choices in our \ourmethod module here.  We conduct the experiments on the channel-wise `hybrid' model and the channel-shared `dynamic' model. Two normalization choices are considered: `softmax' and `sigmoid+softmax'. It is clear that the latter normalization works better (Table~\ref{tab:normalization}). It may boil down to
the nonlinearity introduced by the sigmoid function.

\begin{figure*}
    \centering
    \includegraphics[width=0.999\linewidth]{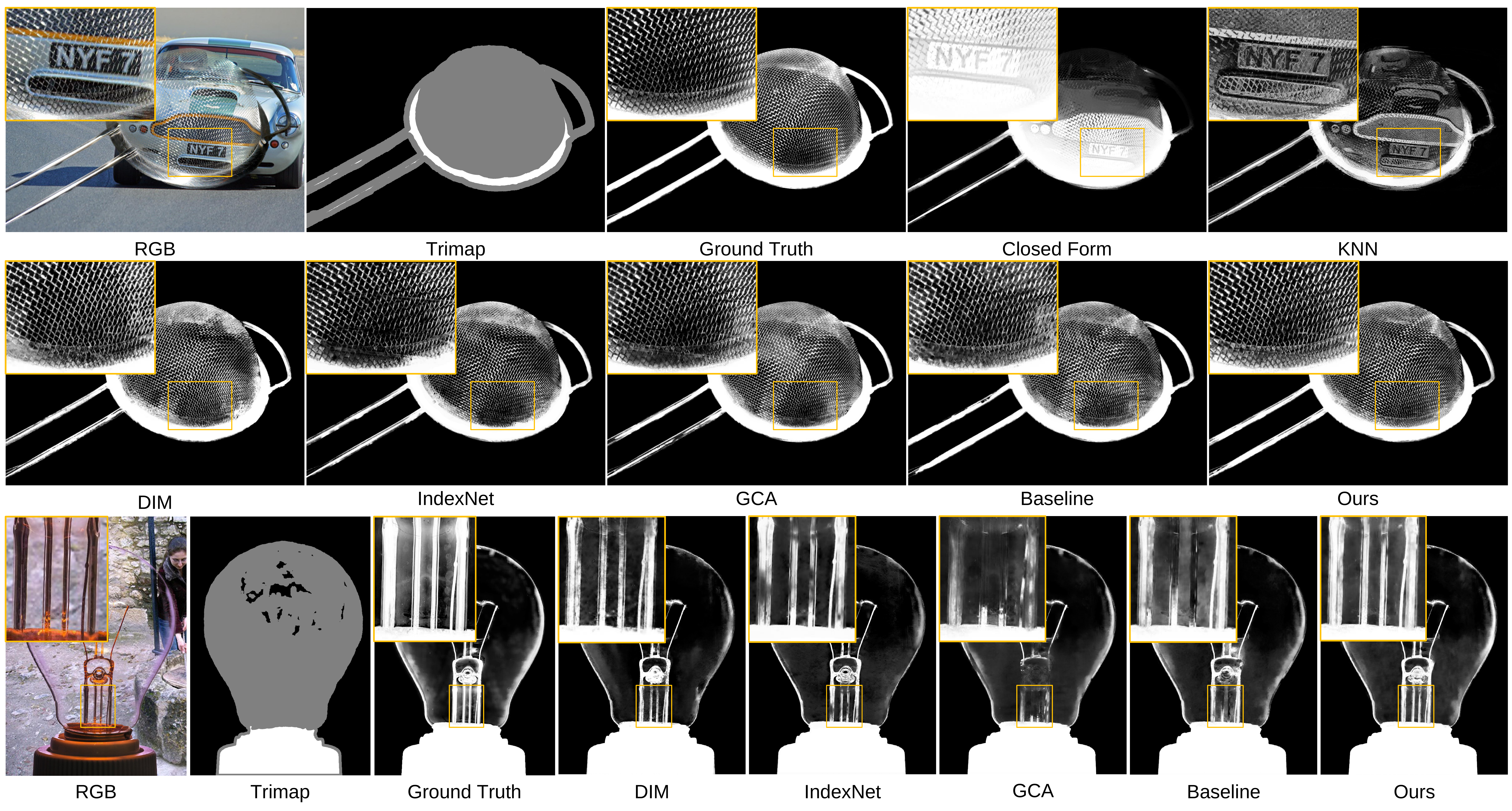}
    \caption{Qualitative results on the Composition-1k test set. The methods in comparison include Closed-Form Matting~\cite{levin2007closed}, KNN Matting~\cite{chen2013knn}, Deep Image Matting (DIM)~\cite{xu2017deep}, IndexNet Matting~\cite{lu2019indices}, GCA Matting~\cite{li2020natural}, our baseline, and our method.}\label{fig:adobe_results}
\end{figure*}

\begin{table}[] \footnotesize
    \centering
    \addtolength{\tabcolsep}{-2.25pt}
    \begin{tabular}{l c c c c c}
    \hline
    Method & Norm & SAD & MSE & Grad & Conn \\
    \hline
       \ourmethod(hybrid-cw) & softmax & 35.93 & 0.0092 & 17.13 & 33.87 \\
       \ourmethod(hybrid-cw)  & sigmoid+softmax & 34.76 & 0.0088 & 16.39 & 32.29  \\
       \ourmethod(dynamic-cs) & softmax & 36.40 & 0.0100 & 17.67 & 34.33 \\
       \ourmethod(dynamic-cs) & sigmoid+softmax & 35.86 & 0.0095 & 17.13 & 33.71 \\
       \hline
    \end{tabular}
\caption{Ablation study of normalization on the Composition-1k test set.}
    \label{tab:normalization}
    \vspace{-10pt}
\end{table}
\vspace{5pt}
\noindent\textbf{Comparison with State of the Art.}
Here we compare our models against other state-of-the-art methods on the Composition-1k test set. Results are shown in Table~\ref{tab:adobe_benchmark}. We observe that our models outperform other methods on all the evaluation metrics with the minimum model capacity. Compared with the state-of-the-art method~\cite{li2020natural}, our best model achieves $8\%$ higher performance with only $40\%$ model
complexity. Our model is also memory-efficient, being able to infer high-resolution images on a single 1080Ti GPU without downsampling on the Composition-1k test set. Some qualitative results are shown in Fig.~\ref{fig:adobe_results}. Our results show improved detail delineation such as the net structure and the filament.
\begin{table}[] \footnotesize
    \centering
    \addtolength{\tabcolsep}{-2pt}
    \begin{tabular}{l c c c c r}
    \hline
      Method  & SAD & MSE & Grad & Conn & \# Params \\
      \hline
      Closed-Form~\cite{levin2007closed} & 168.1 & 0.091 & 126.9 & 167.9 & - \\
      KNN Matting~\cite{chen2013knn} & 175.4 & 0.103 & 124.1 & 176.4 & - \\
       Deep Matting~\cite{xu2017deep} & 50.4 & 0.014 & 31.0 & 50.8 & $>130.55$M \\
       IndexNet Matting~\cite{lu2019indices} & 45.8 & 0.013 & 25.9 & 43.7 & 8.15M \\
       AdaMatting~\cite{cai2019disentangled} & 41.7 & 0.010 & 16.8 & - & - \\
       Context-Aware~\cite{hou2019context} & 35.8 & \textbf{0.0082} & 17.3 & 33.2 & 107.5M \\
       GCA Matting~\cite{li2020natural} & 35.28 & 0.0091 & 16.9 & 32.5 & 25.27M \\
       \hline
        \ourmethod (hybrid-cw)  & 34.76 & 0.0088 & \textbf{16.39} & 32.29 & 8.09M \\
         \ourmethod (dynamic-cs) & 35.86 & 0.0095 & 17.13 & 33.71 & 8.07M \\
\ourmethod (dynamic-cs-d) & \textbf{32.15} & \textbf{0.0082} & \textbf{16.39} & \textbf{29.25} & 8.09M \\
       \hline
    \end{tabular}
\caption{Benchmark results on Composition-1k test set. The best performance is in boldface.}
    \label{tab:adobe_benchmark}
    \vspace{-20pt}
\end{table}

\subsection{The $ \tt alphamatting.com$ Benchmark}
Here we report results on the $\tt alphamatting.com$ online benchmark~\cite{rhemann2009perceptually}. We follow~\cite{li2020natural} to train our model with all the data in the Adobe matting dataset and then test it on the benchmark. As shown in Table~\ref{tab:alphamatting_benchmark}, our method ranks the first w.r.t.\ the gradient error among all published methods. We also achieve comparable overall ranking compared with AdaMatting~\cite{cai2019disentangled} under the SAD and MSE metrics,
suggesting our method is one of the top performing methods on this benchmark.

\begin{table*}[!h]\scriptsize
    \centering
    \addtolength{\tabcolsep}{-3.5pt}
	\renewcommand\arraystretch{1.0}
    \begin{tabular}{l|c|ccc|ccc|ccc|ccc|ccc|ccc|ccc|ccc|ccc}
    \hline
      \multirow{2}{*}{Gradient Error} & \multicolumn{4}{c|}{Average Rank} & \multicolumn{3}{c|}{Troll} & \multicolumn{3}{c|}{Doll} & \multicolumn{3}{c|}{Donkey} & \multicolumn{3}{c|}{Elephant} & \multicolumn{3}{c|}{Plant} & \multicolumn{3}{c|}{Pineapple} & \multicolumn{3}{c|}{Plastic bag} & \multicolumn{3}{c}{Net}  \\
      & Overall & S & L & U  & S & L & U & S & L & U & S & L & U & S & L & U & S & L & U & S & L & U & S & L & U & S & L & U \\
      \hline
      Ours & \textbf{6.3} & 5.6 & \textbf{3.3} & 10.1 & 0.2 & 0.2 & 0.2 & 0.1 & \textbf{0.1} & 0.2 & \textbf{0.1} & \textbf{0.2} & \textbf{0.2} & 0.2 & 0.2 & 0.4 & 1.1 & \textbf{1.3} & 1.9 & 0.6 & 0.7 & 1.7 & 0.6 & \textbf{0.6} & 0.6 & \textbf{0.3} & \textbf{0.3} & 0.4 \\
      \rowcolor{GRAY}
      AdaMatting~\cite{cai2019disentangled} & 7.8 & \textbf{4.5} & 5.6 & 13.3 & 0.2 & 0.2 & 0.2 & \textbf{0.1} & 0.1 & 0.4 & 0.2 & 0.2 & 0.2 & \textbf{0.1} & \textbf{0.1} & 0.3 & \textbf{1.1} & 1.4 & 2.3 & \textbf{0.4} & \textbf{0.6} & \textbf{0.9} & 0.9 & 1 & 0.9 & 0.3 & 0.4 & \textbf{0.4} \\
      GCA Matting~\cite{li2020natural} & 8 & 8.4 & 6.6 & \textbf{9.1} & \textbf{0.1} & \textbf{0.1} & 0.2 & 0.1 & 0.1 & 0.3 & 0.2 & 0.2 & 0.2 & 0.2 & 0.2 & \textbf{0.3} & 1.3 & 1.6 & 1.9 & 0.7 & 0.8 & 1.4 & \textbf{0.6} & 0.7 & \textbf{0.6} & 0.4 & 0.4 & 0.4 \\
      \rowcolor{GRAY}
      Context-aware Matting~\cite{hou2019context} & 9.1 & 10.8 & 9.8 & 6.8 & 0.2 & 0.2 & \textbf{0.2} & 0.1 & 0.2 & \textbf{0.2} & 0.2 & 0.2 & 0.2 & 0.2 & 0.4 & 0.4 & 1.4 & 1.5 & \textbf{1.8} & 0.8 & 1.3 & 1 & 1.1 & 1.1 & 0.9 & 0.4 & 0.4 & 0.4 \\
    \hline
    \end{tabular}
    \vspace{3pt}
    \caption{Gradient errors on the $\tt alphamatting.com$ test set. The top-4 methods are shown. The lowest errors are in boldface.}
    \vspace{-20pt}
    \label{tab:alphamatting_benchmark}
\end{table*}
\subsection{The Distinction-646 Dataset}
We also evaluate our method on the recent
Distinction-646 test set. In Table~\ref{tab:distinction_benchmark}, we report results of the three models performing the best on the Composition-1k dataset and also compare with other benchmarking results provided by~\cite{qiao2020attention}. We have two observations: 1) our models show improved performance against the baseline, which further confirms the effectiveness of our \ourmethod; 2) Our models outperform other reported benchmarking results by large margins, setting a new state of the art on this dataset.

\begin{table}[!t] \footnotesize
    \centering
    \addtolength{\tabcolsep}{1.75pt}
    \begin{tabular}{l c c c c}
    \hline
       Method  & SAD & MSE & Grad & Conn \\
       \hline
       Closed-Form~\cite{levin2007closed} & 105.73 & 0.023 & 91.76 & 114.55 \\
       KNN Matting~\cite{chen2013knn} & 116.68 & 0.025 & 103.15 & 121.45 \\
       Deep Matting~\cite{xu2017deep} & 47.56 & 0.009 & 43.29 & 55.90 \\
\hline
        Baseline-Nearest & 25.03 & 0.0106 & 13.85 & 24.41 \\
        \ourmethod (hybrid-cw) & 24.08 & 0.0104 & 13.53 & 23.59 \\
        \ourmethod (dynamic-cs) & 24.55 & 0.0107 & 14.51 & 23.89 \\
        \ourmethod (dynamic-cs-d) & \textbf{23.20} & \textbf{0.0102} & \textbf{12.39} & \textbf{22.20} \\
        \hline
    \end{tabular}
    \vspace{5pt}
    \caption{Benchmark results on the Distinctions-646 test set. The best performance is in boldface.}
    \label{tab:distinction_benchmark}
    \vspace{-25pt}
\end{table}
\subsection{Visualization of Upsampling Kernels}
\begin{figure}
    \centering
    \includegraphics[width=0.998\linewidth]{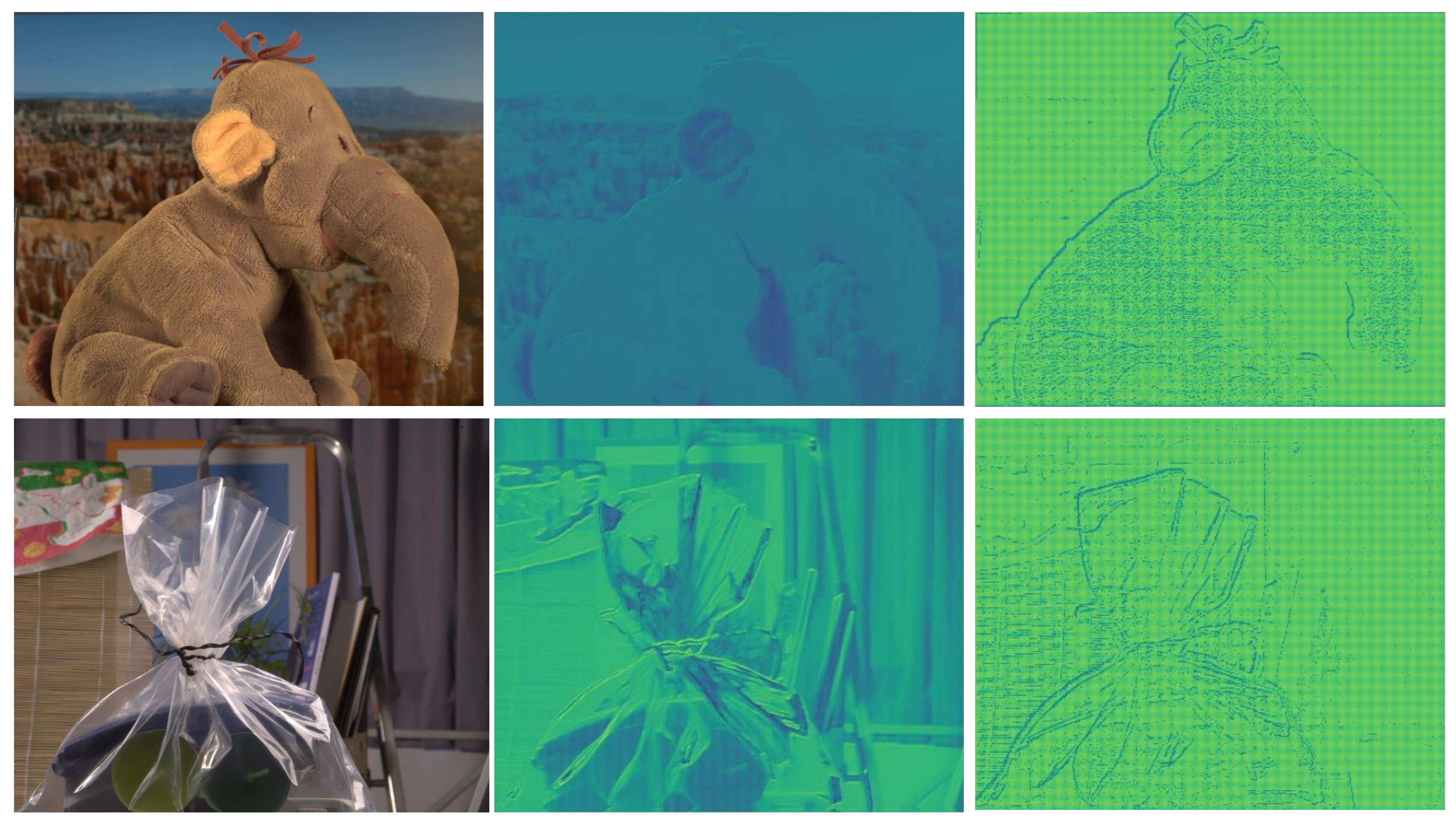}\vspace{-10pt}
    \caption{Visualization of the upsampling kernel. The left is the randomly initialized kernel, and the right is the learned kernel.}
    \label{fig:kernel_visualization}
    \vspace{-15pt}
\end{figure}

Here we visualize the learned upsampling kernel in a `hybrid' model to showcase what is learned by the kernel. Two examples are illustrated in Fig.~\ref{fig:kernel_visualization}. We observe that, after learning, boundary details are highlighted, while flat regions are weakened.

\section{Conclusion}
Considering that affinity is widely exploited in dense prediction, we explore the feasibility to model such second-order information into upsampling for building compact models. We implement this idea with a low-rank bilinear formulation, based on a generalized mathematical view of upsampling. We show that, with negligible parameters increase, our method \ourmethod can achieve better performance on both image reconstruction and image matting tasks. We also investigate different design choices of \ourmethod. Results on three image matting benchmarks all show that \ourmethod achieve
a significant relative improvement and also state-of-the-art results.
In particular, compared with the best performing image matting network,
our model achieves $8\%$ higher performance on the Composition-1k test set, with only $40\%$ model capacity.
For future work, we plan to extend \ourmethod to other dense prediction tasks.

\clearpage
\section*{Appendix}

\appendix
\section{Training Details of Image Reconstruction}
The image reconstruction experiments are implemented on the MNIST dataset~\cite{lecun1998mnist} and Fashion-MNIST dataset~\cite{xiao2017fashion}. They both include $60,000$ training images and $10,000$ test images. During training, the input images are resized to $32\times 32$, and $\ell_1$ loss is used. We use the SGD optimizer with an initial learning rate of $0.01$. The learning rate is decreased by $\times 10$ at the $50$-th, $70$-th, and $85$-th epoch, respectively. We update the parameters for $100$ epochs in total with a batch size of $100$. The evaluation metrics are Peak Signal-to-Noise Ratio (PSNR), Structural SIMilarity (SSIM), Mean Absolute Error (MAE) and root Mean Square Error (MSE).
\vspace{5pt}

\section{Analysis of Complexity}
Here we summarize the model complexity of different implementations of \ourmethod in Table~\ref{tab:complexity_analysis}. We assume that the encoding kernel size is $k\times k$, the upsampling kernel size is $s\times s$, and the channel number of feature map $\mathcal{X}$ is $C$. Since $C$ is much larger than $k$ and $s$, \ourmethod generally has the complexity: $dynamic~cw>hybrid~cw>static~cw>dynamic~cs>hybrid~cs>static~cs$.

\begin{table}[!h]
    \centering
    \begin{tabular}{l|c|c}
    \hline
    Model & Type & \# Params \\
    \hline
       static  & cw & $4\times s\times s+2\times k \times k\times C$ \\
       static  & cs & $4\times s\times s+2\times k \times k$ \\
       hybrid  & cw & $4\times s\times s\times C+2\times k \times k\times C$ \\
       hybrid  & cs & $4\times s\times s\times C+2\times k \times k$ \\
       dynamic  & cw & $4\times s\times s\times C+2\times C \times C$ \\
       dynamic  & cs & $4\times s\times s\times C+2\times C$ \\
      \hline
    \end{tabular}
    \vspace{5pt}
    \caption{Analysis on the complexity of \ourmethod. `cw': channel-wise, `cs': channel-shared}
    \label{tab:complexity_analysis}
\end{table}
\begin{figure*}[!htb]
    \centering
    \includegraphics[width=0.95\linewidth]{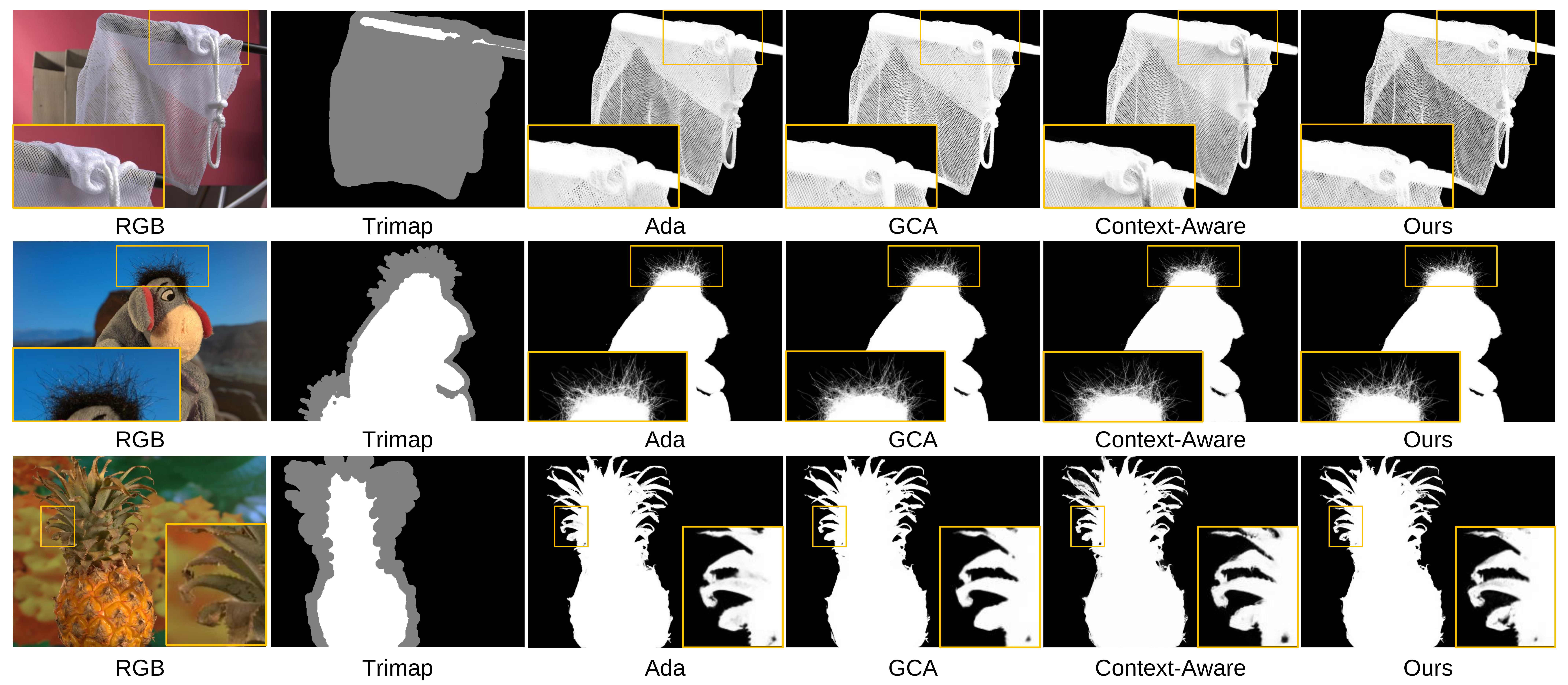}
    \caption{Qualitative results on the \texttt{alphamatting.com} test set. The methods in comparison include AdaMatting~\cite{cai2019disentangled}, GCA Matting~\cite{li2020natural}, Context-Aware Matting~\cite{hou2019context}, and our method.}
    \label{fig:alphamatting}
\end{figure*}

\section{Qualitative Results}
We show additional qualitative results on the \texttt{alphamatting.com} benchmark~\cite{rhemann2009perceptually} in Fig.~\ref{fig:alphamatting}. $4$ top-performing methods are visualized here. Since all these methods achieve good performance, and their quantitative results on the benchmark are very close, it is difficult to tell the obvious difference in Fig.~\ref{fig:alphamatting}. It worth noting that, however, our method produces better visual results on detailed structures, such as gridding of the net, and leaves of the pineapple.

\begin{figure*}[!htb]
    \centering
    \includegraphics[width=0.95\linewidth]{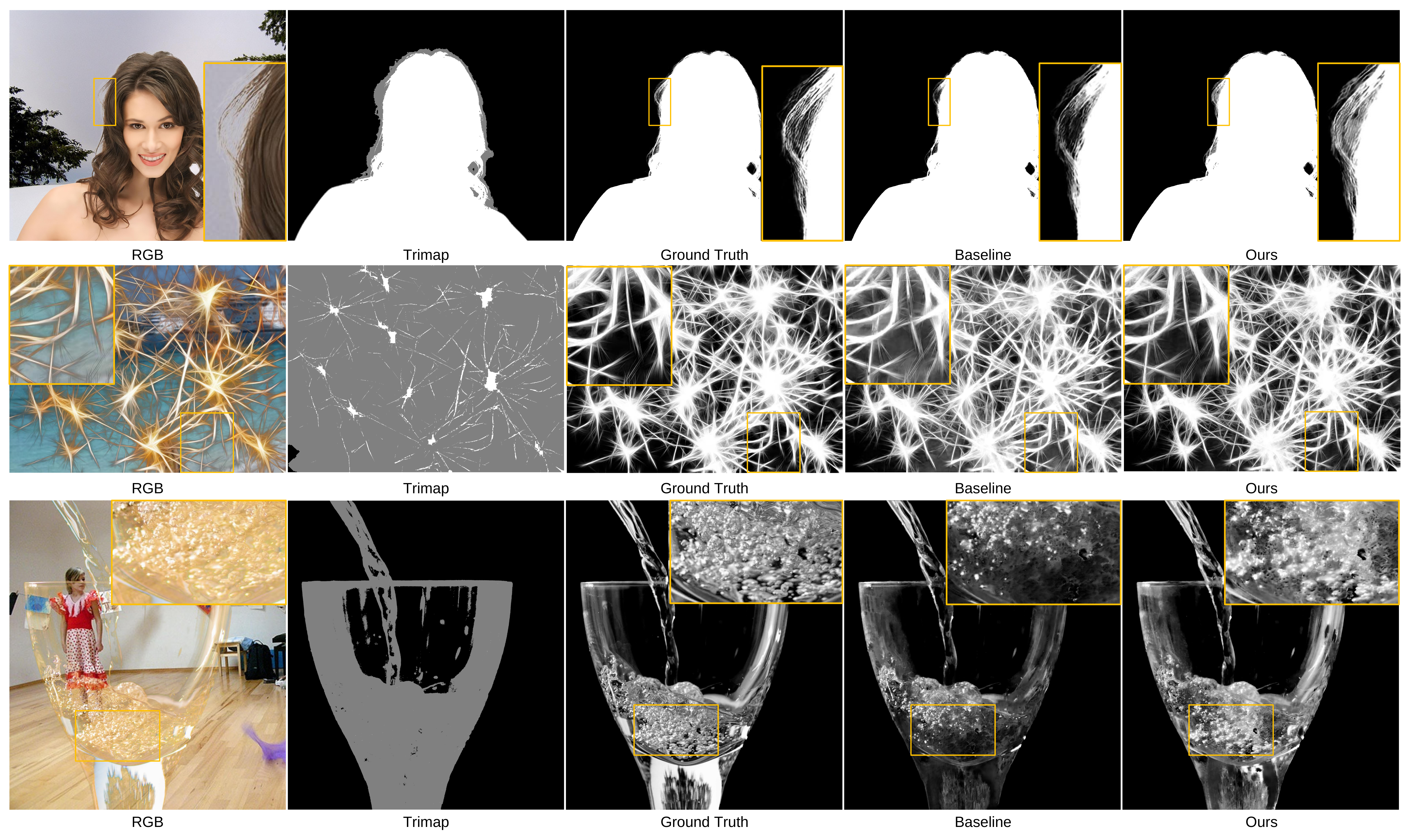}
    \caption{Qualitative results on the Distinction-646 test set. The methods in comparison include the baseline and our method.}
    \label{fig:distinction}
\end{figure*}

We also show qualitative results on the Distinction-646 test set~\cite{qiao2020attention} in Fig.~\ref{fig:distinction}. Since no implementation of other deep methods on this benchmark is publicly available, we only present the results of our baseline and our method here to show the relative improvements. According to Fig.~\ref{fig:distinction}, our method produces clearly
better predictions on highly transparent objects such as the bubbles.

{\small
\bibliographystyle{unsrt}
\bibliography{draft}

\begin{thebibliography}{10}

\bibitem{liu2017learning}
Sifei Liu, Shalini De~Mello, Jinwei Gu, Guangyu Zhong, Ming-Hsuan Yang, and Jan
  Kautz.
\newblock Learning affinity via spatial propagation networks.
\newblock In {\em Advances in Neural Information Processing Systems (NIPS)},
  pages 1520--1530, 2017.

\bibitem{cheng2018depth}
Xinjing Cheng, Peng Wang, and Ruigang Yang.
\newblock Depth estimation via affinity learned with convolutional spatial
  propagation network.
\newblock In {\em Proc. European Conference on Computer Vision (ECCV)}, pages
  103--119, 2018.

\bibitem{gao2019ssap}
Naiyu Gao, Yanhu Shan, Yupei Wang, Xin Zhao, Yinan Yu, Ming Yang, and Kaiqi
  Huang.
\newblock Ssap: Single-shot instance segmentation with affinity pyramid.
\newblock In {\em Proc. IEEE International Conference on Computer Vision
  (ICCV)}, pages 642--651, 2019.

\bibitem{wang2018deep}
Yu~Wang, Yi~Niu, Peiyong Duan, Jianwei Lin, and Yuanjie Zheng.
\newblock Deep propagation based image matting.
\newblock In {\em International Joint Conference on Artificial Intelligence},
  volume~3, pages 999--1006, 2018.

\bibitem{li2020natural}
Yaoyi Li and Hongtao Lu.
\newblock Natural image matting via guided contextual attention.
\newblock In {\em Proc. AAAI Conference on Artificial Intelligence}, volume~34,
  pages 11450--11457, 2020.

\bibitem{lu2019indices}
Hao Lu, Yutong Dai, Chunhua Shen, and Songcen Xu.
\newblock Indices matter: Learning to index for deep image matting.
\newblock In {\em Proc. IEEE International Conference on Computer Vision
  (ICCV)}, pages 3266--3275, 2019.

\bibitem{lu2020index}
Hao Lu, Yutong Dai, Chunhua Shen, and Songcen Xu.
\newblock Index networks.
\newblock {\em IEEE Transactions on Pattern Analysis and Machine Intelligence},
  2020.

\bibitem{wang2019carafe}
Jiaqi Wang, Kai Chen, Rui Xu, Ziwei Liu, Chen~Change Loy, and Dahua Lin.
\newblock Carafe: Content-aware reassembly of features.
\newblock In {\em Proc. IEEE International Conference on Computer Vision
  (ICCV)}, pages 3007--3016, 2019.

\bibitem{shi2016real}
Wenzhe Shi, Jose Caballero, Ferenc Husz{\'a}r, Johannes Totz, Andrew~P Aitken,
  Rob Bishop, Daniel Rueckert, and Zehan Wang.
\newblock Real-time single image and video super-resolution using an efficient
  sub-pixel convolutional neural network.
\newblock In {\em Proc. IEEE Conference on Computer Vision and Pattern
  Recognition (CVPR)}, pages 1874--1883, 2016.

\bibitem{long2015fully}
Jonathan Long, Evan Shelhamer, and Trevor Darrell.
\newblock Fully convolutional networks for semantic segmentation.
\newblock In {\em Proc. IEEE Conference on Computer Vision and Pattern
  Recognition (CVPR)}, pages 3431--3440, 2015.

\bibitem{levin2007closed}
Anat Levin, Dani Lischinski, and Yair Weiss.
\newblock A closed-form solution to natural image matting.
\newblock {\em IEEE Transactions on Pattern Analysis and Machine Intelligence},
  30(2):228--242, 2007.

\bibitem{chen2013knn}
Qifeng Chen, Dingzeyu Li, and Chi-Keung Tang.
\newblock Knn matting.
\newblock {\em IEEE Transactions on Pattern Analysis and Machine Intelligence},
  35(9):2175--2188, 2013.

\bibitem{chuang2001bayesian}
Yung-Yu Chuang, Brian Curless, David~H Salesin, and Richard Szeliski.
\newblock A bayesian approach to digital matting.
\newblock In {\em Proc. IEEE Conference on Computer Vision and Pattern
  Recognition (CVPR)}, volume~2, pages II--II. IEEE, 2001.

\bibitem{he2011global}
Kaiming He, Christoph Rhemann, Carsten Rother, Xiaoou Tang, and Jian Sun.
\newblock A global sampling method for alpha matting.
\newblock In {\em Proc. IEEE Conference on Computer Vision and Pattern
  Recognition (CVPR)}, pages 2049--2056. IEEE, 2011.

\bibitem{cho2016natural}
Donghyeon Cho, Yu-Wing Tai, and Inso Kweon.
\newblock Natural image matting using deep convolutional neural networks.
\newblock In {\em Proc. European Conference on Computer Vision (ECCV)}, pages
  626--643. Springer, 2016.

\bibitem{xu2017deep}
Ning Xu, Brian Price, Scott Cohen, and Thomas Huang.
\newblock Deep image matting.
\newblock In {\em Proc. IEEE Conference on Computer Vision and Pattern
  Recognition (CVPR)}, pages 2970--2979, 2017.

\bibitem{tang2019learning}
Jingwei Tang, Yagiz Aksoy, Cengiz Oztireli, Markus Gross, and Tunc~Ozan Aydin.
\newblock Learning-based sampling for natural image matting.
\newblock In {\em Proc. IEEE Conference on Computer Vision and Pattern
  Recognition (CVPR)}, pages 3055--3063, 2019.

\bibitem{hou2019context}
Qiqi Hou and Feng Liu.
\newblock Context-aware image matting for simultaneous foreground and alpha
  estimation.
\newblock In {\em Proc. IEEE International Conference on Computer Vision
  (ICCV)}, pages 4130--4139, 2019.

\bibitem{cai2019disentangled}
Shaofan Cai, Xiaoshuai Zhang, Haoqiang Fan, Haibin Huang, Jiangyu Liu, Jiaming
  Liu, Jiaying Liu, Jue Wang, and Jian Sun.
\newblock Disentangled image matting.
\newblock In {\em Proc. IEEE International Conference on Computer Vision
  (ICCV)}, pages 8819--8828, 2019.

\bibitem{lutz2018alphagan}
Sebastian Lutz, Konstantinos Amplianitis, and Aljosa Smolic.
\newblock Alphagan: Generative adversarial networks for natural image matting.
\newblock {\em British Machince Vision Conference (BMVC)}, 2018.

\bibitem{badrinarayanan2017segnet}
Vijay Badrinarayanan, Alex Kendall, and Roberto Cipolla.
\newblock {SegNet}: {A} deep convolutional encoder-decoder architecture for
  image segmentation.
\newblock {\em IEEE Transactions on Pattern Analysis and Machine Intelligence},
  39(12):2481--2495, 2017.

\bibitem{kim2016hadamard}
Jin-Hwa Kim, Kyoung-Woon On, Woosang Lim, Jeonghee Kim, Jung-Woo Ha, and
  Byoung-Tak Zhang.
\newblock Hadamard product for low-rank bilinear pooling.
\newblock {\em arXiv preprint arXiv:1610.04325}, 2016.

\bibitem{yu2018hierarchical}
Chaojian Yu, Xinyi Zhao, Qi~Zheng, Peng Zhang, and Xinge You.
\newblock Hierarchical bilinear pooling for fine-grained visual recognition.
\newblock In {\em Proc. European Conference on Computer Vision (ECCV)}, pages
  574--589, 2018.

\bibitem{yang2019condconv}
Brandon Yang, Gabriel Bender, Quoc~V Le, and Jiquan Ngiam.
\newblock Condconv: Conditionally parameterized convolutions for efficient
  inference.
\newblock In {\em Advances in Neural Information Processing Systems (NIPS)},
  pages 1307--1318, 2019.

\bibitem{lecun1998mnist}
Yann LeCun.
\newblock The mnist database of handwritten digits.
\newblock {\em http://yann. lecun. com/exdb/mnist/}, 1998.

\bibitem{xiao2017fashion}
Han Xiao, Kashif Rasul, and Roland Vollgraf.
\newblock Fashion-mnist: a novel image dataset for benchmarking machine
  learning algorithms.
\newblock {\em arXiv preprint arXiv:1708.07747}, 2017.

\bibitem{he2016deep}
Kaiming He, Xiangyu Zhang, Shaoqing Ren, and Jian Sun.
\newblock Deep residual learning for image recognition.
\newblock In {\em Proc. IEEE Conference on Computer Vision and Pattern
  Recognition (CVPR)}, pages 770--778, 2016.

\bibitem{rotabulo2017place}
Samuel Rota~Bul\`o, Lorenzo Porzi, and Peter Kontschieder.
\newblock In-place activated batchnorm for memory-optimized training of dnns.
\newblock In {\em Proc. IEEE Conference on Computer Vision and Pattern
  Recognition (CVPR)}, 2018.

\bibitem{ronneberger2015u}
Olaf Ronneberger, Philipp Fischer, and Thomas Brox.
\newblock U-net: Convolutional networks for biomedical image segmentation.
\newblock In {\em Proc. International Conference on Medical Image Computing and
  Computer-Assisted Intervention (MICCAI)}, pages 234--241. Springer, 2015.

\bibitem{lin2014microsoft}
Tsung-Yi Lin, Michael Maire, Serge Belongie, James Hays, Pietro Perona, Deva
  Ramanan, Piotr Doll{\'a}r, and C~Lawrence Zitnick.
\newblock Microsoft coco: Common objects in context.
\newblock In {\em Proc. European Conference on Computer Vision (ECCV)}, pages
  740--755. Springer, 2014.

\bibitem{everingham2010pascal}
Mark Everingham, Luc Van~Gool, Christopher~KI Williams, John Winn, and Andrew
  Zisserman.
\newblock The pascal visual object classes (voc) challenge.
\newblock {\em International Journal of Computer Vision}, 88(2):303--338, 2010.

\bibitem{rhemann2009perceptually}
Christoph Rhemann, Carsten Rother, Jue Wang, Margrit Gelautz, Pushmeet Kohli,
  and Pamela Rott.
\newblock A perceptually motivated online benchmark for image matting.
\newblock In {\em Proc. IEEE Conference on Computer Vision and Pattern
  Recognition (CVPR)}, pages 1826--1833. IEEE, 2009.

\bibitem{qiao2020attention}
Yu~Qiao, Yuhao Liu, Xin Yang, Dongsheng Zhou, Mingliang Xu, Qiang Zhang, and
  Xiaopeng Wei.
\newblock Attention-guided hierarchical structure aggregation for image
  matting.
\newblock In {\em Proc. IEEE Conference on Computer Vision and Pattern
  Recognition (CVPR)}, pages 13676--13685, 2020.

\bibitem{paszke2019pytorch}
Adam Paszke, Sam Gross, Francisco Massa, Adam Lerer, James Bradbury, Gregory
  Chanan, Trevor Killeen, Zeming Lin, Natalia Gimelshein, Luca Antiga, et~al.
\newblock Pytorch: An imperative style, high-performance deep learning library.
\newblock In {\em Advances in Neural Information Processing Systems (NIPS)},
  pages 8026--8037, 2019.

\bibitem{krizhevsky2017imagenet}
Alex Krizhevsky, Ilya Sutskever, and Geoffrey~E Hinton.
\newblock Imagenet classification with deep convolutional neural networks.
\newblock {\em Communications of the ACM}, 60(6):84--90, 2017.

\bibitem{kingma2014adam}
Diederik~P Kingma and Jimmy Ba.
\newblock Adam: A method for stochastic optimization.
\newblock {\em arXiv preprint arXiv:1412.6980}, 2014.

\end{thebibliography}
}

\end{document}